\documentclass[10pt,twocolumn,letterpaper]{article}

\usepackage{wacv}              % To produce the CAMERA-READY version

\usepackage[accsupp]{axessibility} % Improves PDF readability for those with disabilities.

\usepackage{graphicx}
\usepackage{amsmath}
\usepackage{amssymb}
\usepackage{booktabs}
\usepackage{listings}
\usepackage{multirow}
\usepackage{ctable}
\usepackage{indentfirst}
\usepackage{bbding}
\usepackage{pifont}
\usepackage{wasysym}
\usepackage{amssymb}
\usepackage{enumitem}
\usepackage{newverbs}

\usepackage{bm}
\usepackage{dsfont}
\usepackage{balance}

\usepackage[pagebackref,breaklinks,colorlinks]{hyperref}

\usepackage[capitalize]{cleveref}
\crefname{section}{Sec.}{Secs.}
\Crefname{section}{Section}{Sections}
\Crefname{table}{Table}{Tables}
\crefname{table}{Tab.}{Tabs.}

\usepackage{array}
\newcolumntype{P}[1]{>{\centering\arraybackslash}p{#1}}

\newcommand{\1}[1]{\mathds{1}_{\left(#1\right)}}

\definecolor{c1}{RGB}{228,26,27}
\definecolor{c2}{RGB}{55,126,184}
\definecolor{c3}{RGB}{84,178,81}
\definecolor{c4}{RGB}{156,85,167}
\definecolor{c5}{RGB}{255,153,52}

\newverbcommand{\oneverb}{\color{c1}}{}
\newverbcommand{\twoverb}{\color{c2}}{}
\newverbcommand{\threeverb}{\color{c3}}{}
\newverbcommand{\fourverb}{\color{c4}}{}
\newverbcommand{\fiveverb}{\color{c5}}{}

\definecolor{lg}{RGB}{148,179,175}
\definecolor{mg}{RGB}{84,138,133}
\definecolor{gc}{RGB}{139,207,151}
\definecolor{babyblueeyes}{rgb}{0.63, 0.79, 0.95}
\definecolor{brickred}{rgb}{0.8, 0.25, 0.33}

\def\wacvPaperID{117} % *** Enter the WACV Paper ID here
\def\confName{WACV}
\def\confYear{2024}

\newcommand{\myparagraph}[1]{\smallskip\noindent\textbf{#1.}}

\begin{document}

\title{Semantic-aware Video Representation for Few-shot Action Recognition}

\author{Yutao Tang\\
Johns Hopkins University\\
{\tt\small ytang67@jhu.edu}
\and
Benjam\'in B\'ejar\\
Paul Scherrer Institut\\
{\tt\small benjamin.bejar@psi.ch}
\and
Ren\'e Vidal\\
University of Pennsylvania\\
{\tt\small vidalr@upenn.edu}
}

\maketitle

%%%%%%%%% ABSTRACT
\begin{abstract}
    Recent work on action recognition leverages 3D features and textual information to achieve state-of-the-art performance. However, most of the current few-shot action recognition methods still rely on 2D frame-level representations, often require additional components to model temporal relations, and employ complex distance functions to achieve accurate alignment of these representations. In addition, existing methods struggle to effectively integrate textual semantics, some resorting to concatenation or addition of textual and visual features, and some using text merely as an additional supervision without truly achieving feature fusion and information transfer from different modalities. In this work, we propose a simple yet effective \textbf{S}emantic-\textbf{A}ware \textbf{F}ew-\textbf{S}hot \textbf{A}ction \textbf{R}ecognition (\textbf{SAFSAR}) model to address these issues. We show that directly leveraging a 3D feature extractor combined with an effective feature-fusion scheme, and a simple cosine similarity for classification can yield better performance without the need of extra components for temporal modeling or complex distance functions. We introduce an innovative scheme to encode the textual semantics into the video representation which adaptively fuses features from text and video, and encourages the visual encoder to extract more semantically consistent features. In this scheme, SAFSAR achieves alignment and fusion in a compact way. Experiments on five challenging few-shot action recognition benchmarks under various settings demonstrate that the proposed SAFSAR model significantly improves the state-of-the-art performance.

\end{abstract}

%%%%%%%%% BODY TEXT
\section{Introduction}
\label{sec:intro}
    Few-shot action recognition (FSAR) is a challenging and practical problem in the field of computer vision. Unlike conventional action recognition which typically relies on large amounts of annotated data, and where the goal at inference time is to correctly classify actions that have been seen during training, FSAR aims to develop models capable of accurately identifying unseen action categories through just a handful of annotated samples per category. Consequently, FSAR is a promising alternative to conventional action recognition when collecting abundant annotated data for each action category is impractical or very costly~\cite{zhu2018cmn}.
    % This task is particularly relevant in real-world scenarios where collecting abundant annotated data for each action category is impractical or costly~\cite{zhu2018cmn}. 
    Specifically, an FSAR problem involves a support set comprised of a few labelled videos and a corresponding query set which contains unlabelled videos. The goal of FSAR is to assign a correct label to query videos merely using those few labelled videos in the support set as reference.
    
    Mainstream approaches to FSAR \cite{zhu2018cmn, bishay2019tarn, cao2020otam, zhang2020arn, zhang2021itanet, perrett2021trx, wang2022hyrsm, zhang2023mastaf}  rely on metric-based meta-learning  \cite{snell2017proto}. Such approaches first learn to generate representations for both support and query videos, and then learn a similarity metric to associate query videos with one of the categories in the support set, akin to the nearest neighbor classifier. For example, OTAM \cite{cao2020otam} extracts frame-level representations and designs an ordered temporal alignment distance function to measure the similarity between support and query videos. TRX \cite{perrett2021trx} exploits the CrossTransformer \cite{doersch2020xtf} to highlight query-related sub-sequences of support videos and constructs tuples of varying number of frames to compare support and query videos. HyRSM \cite{wang2022hyrsm} proposes a hybrid relation module that explores within- and cross-video relations to extract task-specific representations and designs a set matching metric to get the query-support correspondence.
    
    % ARN \cite{zhang2020arn} uses 3D-CNN with temporal and spatial attention to extract discriminative spatio-temporal representations and adopts the style of Relation Network \cite{sung2018rn} to quantify the relation between support and query videos.
    
    % limitation 1: (a) 2d representation, temporal relation is modelded by using additional module or encoded in the distance function (e.g. OTAM, measuring temporal alignment). (b) complicated distance func demands more computations
    Despite recent advances in the field, current approaches still have two considerable limitations. First, most of the state-of-the-art (SoTA) methods rely on 2D feature extractors to obtain frame-level representations, employ additional components to model temporal dependencies and enforce temporal alignment with a complex distance function. While this type of approaches achieve considerable classification results, the advances in general video action recognition indicate that 3D feature extractors, such as 3D-CNNs \cite{carreira2017kinetics, feichtenhofer2019slowfast, hara2017restnet3d} and video transformers \cite{arnab2021vivit, bertasius2021timesformer, tong2022vmae}, hold significant promise in capturing spatio-temporal features more accurately. This is particularly evident during action sequences characterized by intricate temporal movements, where these advanced models demonstrate their potential to excel. In light of these advances, it is beneficial to leverage 3D feature extractors, avoiding the need of extra components or intricate distance functions in the FSAR task.
%    {\color{brickred} \sout{Furthermore, while adopting less computationally expensive 2D feature extractors might seem appealing at first glance, it might actually reduce the overall effectiveness because of the extra computational demand via the introduction of intricate distance functions for finding the matches between query and support examples. For instance, TRX \cite{perrett2021trx} considers all the sub-sequences of varying number of frames to represent videos for comparison during matching, significantly amplifying the computational costs as the number of sampled frames increases.}}
    % limitation 2: no semantic
    Additionally, the second important limitation of current approaches for FSAR concerns the integration and fusion of textual information with visual cues. 
    %%% 说这里太strong，rene说可以这样：说有哪些combine的technique，很好很好，blabla，但现在的methods却没有用
    % Previous attempts remain insufficient due to simplistic fusion strategies or neglect of fusion. 
    % a summary of multi-modality fusion?
    Previous attempts struggle to fully leverage the textual descriptors due to simplistic fusion strategies \cite{zhang2020mem, wang2021semguide} or by treating text as merely an additional supervision without truly fusing features from different modalities \cite{jin2023eant, shi2022knowledge}.
    For example, \cite{zhang2020mem} employs an LSTM-alike memory network to get dynamic visual and textual features but then simply concatenate them as multi-modal features for few-shot recognition. \cite{wang2021semguide} learns a generative model to project visual features into the textual semantic space, and sums the visual features and the generated semantic features element wise for few-shot classification. \cite{ni2022morn} fuses visual and textual features through weighted average. \cite{jin2023eant} treats text as a supplemental supervision to regularize the visual features of the query video without truly fusing the textual features with the visual features. Besides, these methods do not harness the full potential of the pre-trained language models as they use Word2Vec \cite{mikolov2013word2vec} embeddings or handcrafted sentence templates, \eg \textit{a video of \{action\}}. These shortcomings substantially hinder effective integration of textual semantics.

    % most of them use text as an additional supervisiory signal and the text-related module is presented as a standalone module in addition to the main architecture. 

    % align-then-fuse, 但我们是 fuse-then-align，对metric learning更好!
    % 他们把align和fuse分裂搞，但我们是unify地搞

    % besides, 他们用CLIP就是 a video of xxx is playing。这并没有真的fully harness the power of the language model。inspired by \cite{rehears}，我们用他们的elaborative descriptions for UCF101 and HMDB51, better 

    %% 这意思是可以帮助visual model更好地focus在text说的动作上！
    % prompt learning to guide the learning process by helping models better focus on the descriptions or
    % instructions associated with actions in the input videos

    In this paper, we address the aforementioned two limitations by proposing a novel \textbf{S}emantic-\textbf{A}ware \textbf{F}ew-\textbf{S}hot \textbf{A}ction \textbf{R}ecognition (\textbf{SAFSAR}) model. Specifically, we leverage recent advances in general action recognition and use a 3D feature extractor to directly extract a discriminative spatio-temporal representation for both support and query videos. 
    % Unlike previous approaches relying on separate modules for handling temporal relations, SAFSAR streamlines its architecture for simplicity, efficiency, and applicability to various settings. 
    Further, we design a novel paradigm that achieves seamless integration of the textual semantics by aligning and fusing the visual and textual features in a compact way. Specifically, we first develop a lightweight fusion module to adaptively encode textual semantics into the video representations of the support set while the query video representations remain untouched. Next, we apply a task-specific learning module to enrich the video representations with the interactions across the support and query videos. Finally, we classify the query video using a cosine similarity distance between the final representations of the support and query videos. This paradigm design provides a natural mechanism for textual semantics and visual features to be learned and fused in a shared latent knowledge space because we are aligning semantic-unaware query representations with semantic-aware support representations. By this means, we not only effectively fuse features from the two modalities but also encourage the visual encoder to extract more semantically consistent features.
    
    % %% text
    % This model design provides a
    % natural mechanism for visual and semantic representations
    % to be learned in a shared knowledge space, which can bridge
    % the semantic gap and encourage the learned visual embedding to be discriminative and more semantically consistent.
    
    In summary, our contributions are three-fold. (1) We propose a simple yet effective multi-modal model (SAFSAR) for few-shot action recognition that extracts discriminative spatio-temporal features while simultaneously exploiting textual semantics. (2) We design a novel paradigm for aligning and fusing visual features and textual semantics in a compact way. (3) We perform extensive evaluations on five challenging benchmarks to demonstrate the superiority of our approach over prior art.
    % \begin{enumerate}[leftmargin=*]
    % %[label={(\arabic*)}]%[leftmargin=0.8cm]
    
%     \item 
    
%     \item 
    
%     \item 
% % \end{enumerate}

% ======================================= model figure
\begin{figure*}
    \centering
    \setlength{\belowcaptionskip}{-0.4cm} 

    \includegraphics[trim=0cm 0cm 0cm 0cm, clip, width=\textwidth]{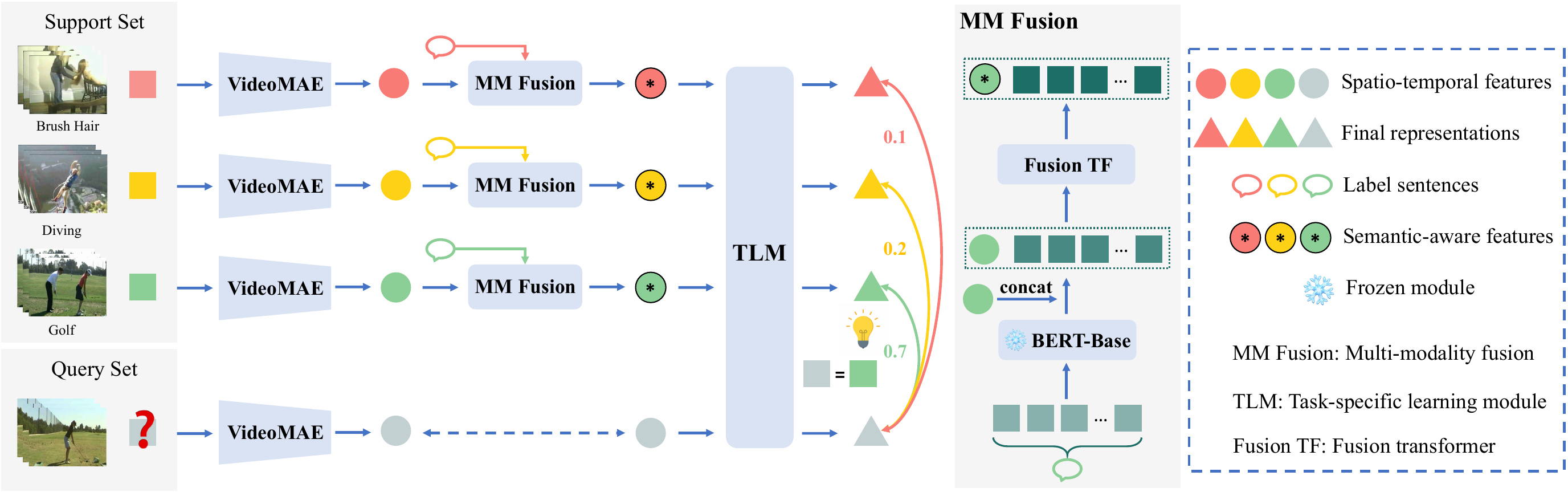}
    
    \caption{\textbf{Left:} Model architecture of our proposed SAFSAR illustrated in a 3-way 1-shot problem. We first extract spatio-temporal features from the support and query videos using VideoMAE. Then, the support video features and their corresponding label sentences are input to the multi-modality fusion (MM Fusion) module to obtain semantic-aware support features, while the query video features remain untouched. Finally, the support and query features are input to a task-specific learning module (TLM) to get the final representations which are then used to compute the cosine similarity scores between the query and support videos. Subsequently, we apply softmax to the cosine similarity scores, obtaining the probabilities of classifying the query video against support action classes.
    \textbf{Middle:} Details of the MM Fusion module. The label description sentence is first tokenized 
    % (shown in the bottom as \lightgrn{light green squares})
    and then forwarded through the frozen BERT-Base model, obtaining textual features for each token. 
    % (shown in the middle as \mediumgrn{medium green squares})
    Then, we concatenate the support features
    % (shown in the middle as a \grncircle{green circle})
    as an extra token to the textual feature tokens, which are then forwarded through a fusion transformer (Fusion TF) to obtain the semantic-aware support features.
    % (shown in the top as a black outlined \grncircle{green circle} with a star in the middle)
    \textbf{Right:} The legend of this figure.}
    
    \label{fig:model}
\end{figure*}
% ======================================= model figure

%-------------------------------------------------------------------------
\section{Related Work}

% \subsection{Few-shot Learning}
\myparagraph{Few-shot Learning}
    Few-shot learning approaches can be broadly divided into three categories: data-level, optimization-level, and metric-level approaches. The data-level approaches \cite{dhillon2019, wang2020, chen2019, ratner2017} address the few-shot learning problem by enlarging the datasets through collection or synthesis of data to enhance model generalization for tasks with scarce data. They accomplish this using pre-training and fine-tuning techniques. The optimization-level approaches \cite{andrychowicz2016, antoniou2018, finn2017maml, ravi2017, zhang2020, nichol2018reptile} learn an optimized model that can serve as a good initialization and that can be quickly adapted to a new task given only limited amount of data. The metric-level approaches concentrate on developing a latent feature embedding space where the similarities between the query set and the support set representations can be measured effectively. For example, Matching Network \cite{vinyals2016mn} is among the first to implement the latter idea for one-shot learning. ProtoNet \cite{snell2017proto} extends it and proposes to use the average embeddings as the prototype of each class in the few-shot setting and uses the Euclidean distance instead of cosine similarity to compare the embeddings. Relation Network \cite{sung2018rn} further improves it by employing a learned distance function. Our work is a metric-level approach that shares the same spirit of ``learn-to-compare" and we focus on the more challenging few-shot action recognition task which involves diverse spatio-temporal dependencies.

% \subsection{Unimodal FSAR}
\myparagraph{Unimodal FSAR}
    Few-shot action recognition (FSAR) is a sub-field of few-shot learning and it deals with complex high-dimensional videos. Existing techniques for FSAR generally follow the metric-level approaches, typically using a single modality, \ie the RGB videos alone. In terms of feature extraction, most methods use 2D feature extractors, like ResNet-50 \cite{he2016resnet}, to extract frame-level representations. 
    % 2D
    CMN \cite{zhu2018cmn} uses a multi-saliency embedding algorithm to selectively combine the frame-level representations into a video representation. OTAM \cite{cao2020otam} compares the support and query videos using an ordered temporal alignment distance function on the frame-level representations. ITANet \cite{zhang2021itanet} proposes a frame-wise implicit temporal alignment network to model the temporal context from the frame-level representations. TRX \cite{perrett2021trx} exploits CrossTransformer \cite{doersch2020xtf} to highlight the query-related sub-sequences of support videos and matches videos through plentiful tuples of varying number of frames. STRM \cite{thatipelli2022strm} applies spatio-temporal enrichment to the frame-level representations and uses TRX with a single cardinality to classify the query video. HyRSM \cite{wang2022hyrsm} proposes a hybrid relation module that models within- and cross-video relations to extract task-specific representations and a bidirectional Mean Hausdorff Metric (Bi-MHM) to get the query-support correspondence.
    % 3D: protogan, generalized... 这两篇都是给generalized，可能可以不写？
    % 3D: ARN(permutation-invariant), kumar2019protogan, tarn
    Besides, very few approaches attempt at incorporating 3D feature extractors to directly extract spatio-temporal representations. TARN \cite{bishay2019tarn} uses C3D \cite{tran2015c3d} to extract snippet features and apply snippet-based attention to achieve temporal alignment with a deep-distance metric. ARN \cite{zhang2020arn} also uses C3D \cite{tran2015c3d} for capturing short-range temporal context and introduces permutation-invariant pooling with spatial and temporal attention to obtain robust descriptors for videos. However, these approaches have inferior performance to recent methods using 2D features \cite{wang2022hyrsm, thatipelli2022strm}. Moreover, the advances on the video transformers \cite{tong2022vmae, arnab2021vivit, bertasius2021timesformer, liu2022videoswintf} suggest its promising application in FSAR. MASTAF \cite{zhang2023mastaf} proposes a model-agnostic network with self- and cross-attention to classify the query video. They demonstrate using ViViT \cite{arnab2021vivit} achieves the best results compared to using 2D- or 3D-CNNs. However, they only explore 1-shot setting using ViViT.
    Our work leverages VideoMAE \cite{tong2022vmae}, a Transformer-based \cite{vaswani2017tf} model that is capable of capturing intricate spatio-temporal relationships within videos. We explore 1-shot, 3-shot, and 5-shot settings and achieve superior performance than MASTAF \cite{zhang2023mastaf} as shown in \cref{tab:sota}.
    
% \subsection{Multi-modal FSAR}
\myparagraph{Multi-modal FSAR}
    Some recent work has explored the inclusion of additional modalities such as motion \cite{wanyan2023active, wang2023molo, wu2022motion}, depth \cite{fu2020amefu}, object features \cite{huang2022compound} and text \cite{zhang2020mem, wang2021semguide, shi2022knowledge, ni2022morn, wang2023dual, jin2023eant}. Of particular interest are those approaches that explore the use of textual information in conjunction with visual cues for improved FSAR.
    \cite{zhang2020mem} proposes a memory network to extract visual and textual features dynamically and then fuses them by concatenation for further recognition. \cite{wang2021semguide} trains a generative model to imitate the semantic labels and fuses the visual features and the generated textual features by element-wise addition. \cite{shi2022knowledge} leverages the pre-trained vision-language model to generate matching scores between video frames and handcrafted action-related phrases. These scores are subsequently used as descriptors for each video to train a temporal network for action classification. \cite{ni2022morn} computes prototypes for both modalities respectively and then fuses them through a weighted average. \cite{wang2023dual} obtains enhanced visual features by channel-wise multiplication of the visual vectors and the semantic vectors. \cite{jin2023eant} uses semantic labels as a supplemental supervision to regularize the visual features of the query video without truly fusing the textual features with the visual features. While promising progress has been made, these approaches fall short of fully integrating textual and visual features because either they resort to simplistic fusion strategies, or they simply treat text as an auxiliary source of guidance instead of truly integrating it with visual features. This gap in the literature inspired us to design a novel paradigm in SAFSAR to combine multi-modal features seamlessly.

%-------------------------------------------------------------------------
\section{Methods}
\label{sec:methods}

\subsection{Few-Shot Action Recognition}
We consider the following setup for multi-modal FSAR where we have access to a training set of triplets $\big\{(y_i,\bm x_i, \tau_i) \big\}_{i=1}^n$, where $y_i\in\{1,\ldots,C\}$ is the action class label, $\bm x_i$ represents a video clip of $T_i$ frames, and $\tau_i$ is a text description of the activity. In the few-shot learning paradigm, a subset of labeled videos, \ie the {\em support set} is used to classify a novel unlabeled {\em query} video. We follow the $N$-way $K$-shot approach which means that the support set contains a subset of videos that represent $N\leq C$ of the total number of classes, and with $K$ instances per class. The query video has an unknown class label belonging to one of the classes in the support set. To make the training more consistent with the testing scenario, we adopt episodic training as in \cite{cao2020otam,perrett2021trx,wang2022hyrsm}. Specifically, at each training episode $e\in{1,\ldots,E}$, where $E$ is the number of episodes, we extract a support set $\mathcal{S}_e \subset \{1,\ldots,n\}$ as follows: We first randomly sample a subset of $N$ classes out of the  total number of classes $C$ and for each class we sample $K$ instances at random without replacement. This means that the support set contains a total of $N\times K$ triplets from the training set. Next, we sample a query set $\mathcal{Q} = \{q_e\}$ consisting of a single video from one of the classes in the support set, where $q_e$ is sampled at random from $\{1,\ldots,n\}\backslash\mathcal{S}_e$. A classifier is then trained using all episodes.

\begin{table*}[htbp]
  \centering
  \resizebox{\textwidth}{!}{

    \begin{tabular}{lccc|rrr|rrr|rrr}

    \toprule[1pt]
    
    \multicolumn{1}{c}{\multirow{2}[0]{*}{Methods}} & \multicolumn{3}{c}{SSv2-Full} & \multicolumn{3}{c}{SSv2-Small} & \multicolumn{3}{c}{UCF101} & \multicolumn{3}{c}{HMDB51} \\

    \cmidrule[1pt]{2-13}
    
  & 1-shot     & 3-shot     & 5-shot     & 1-shot     & 3-shot     & 5-shot     & 1-shot     & 3-shot     & 5-shot     & 1-shot     & 3-shot     & 5-shot \\
    
    \midrule[1pt] 
    \hspace{0.4cm} \textit{Unimodal FSAR} \\
    
    ARN \cite{zhang2020arn} & -    & -   & -   & -  & -  & -  & 66.30  & -  & 83.10  & 45.50  & - & 60.60 \\
    OTAM \cite{cao2020otam} & 42.80    & 51.50  & 52.30   & 36.40  & 45.90  &  48.00  & 79.90  & 87.00  & 88.90  & 54.50  & 65.70 & 68.00 \\
    CMN-J \cite{zhu2020cmnj} &   -    &    -   & -  &  36.20     &    44.60   &    48.80   &   -    &  -     & -  &     -  &     -  & - \\
    ITANet \cite{zhang2021itanet} & 49.20    & 59.10   & 62.30   & 39.80  & 49.40  &  53.70  & -  & -  & -  & -  & - & -\\
    TRX ($\Omega$={1}) \cite{perrett2021trx} & 38.80    & 54.40  & 60.60   & 34.90   &   & 47.60  &  53.30  & -  & -  & -  & -  & - \\
    TRX ($\Omega$={2,3}) \cite{perrett2021trx}   & 42.00    & 57.60  & 64.60  & 36.00    & 51.90  & 59.10  & 78.20  & 92.40  & 96.10  & 53.10  & 66.80  & 75.60 \\
    T${\text{A}^2}$N \cite{li2022ta2n} &   47.60    &    -   & 61.00  &  -     &    -   &    -   &   81.90    &  -     & 95.10 &     59.70  &     -  & 73.90 \\
    STRM \cite{thatipelli2022strm}  &   -    &    -   & 70.20  &  -     &    -   &    -   &   -    &  -     & 98.10  &     -  &     -  & 81.30 \\
    MTFAN \cite{wu2022motion}  &   45.70  &    -   & 60.40 &  -     &    -   &    -   &   84.80   &  -     & 95.10  &     59.00  &     -  & 74.60 \\
    HyRSM \cite{wang2022hyrsm} & 54.30  & \underline{65.10}  & 69.00    & 40.60  & \underline{52.30}  & 56.10  & 83.90  & 93.00    & 94.70  & 60.30  & 71.70  & 76.00 \\
    TRX+L2A \cite{gowda2022l2a} &    -   &    -   &   - &    -   &    -   &   - &  79.20  & 93.20  & 96.30 &  51.90  & 68.20  & 77.00 \\
    ATA \cite{nguyen2022inductive} &   43.80    &    -   & 61.10  &  -     &    -   &    -   &   84.90   &  -     & 95.90  &     59.60  &     -  & 76.90 \\
    HCL \cite{zheng2022hcl} &   47.30    &    59.00   & 64.90  &  38.70     &    49.10   &    55.40   &   82.50    &  91.00    & 93.90  &   59.10 &   71.20  & 76.30  \\
    CPM \cite{huang2022compound} &   49.30    &    -   & 66.70  &  38.90    &    -   &    61.60   &   71.40    &  -     & 91.00  &     60.10  &     -  & 77.00 \\
    MASTAF (TSN) \cite{zhang2023mastaf} & 46.90  &   -    &    62.40   & 37.50  &   -    &  50.20    & 79.30  &   -    &  90.3    & 54.80  &   -    & 67.70 \\
    MASTAF (R3D) \cite{zhang2023mastaf} & 50.30  &   -    &    66.70   & 39.90  &   -    &  52.20    & 90.60  &   -    &  97.60    & 67.90  &   -    & 81.20 \\
    MASTAF (ViViT) \cite{zhang2023mastaf} & \underline{60.70}  &   -    &    -   & \underline{45.60}  &   -    &  -     & \underline{91.60}  &   -    &  -     & \underline{69.50}  &   -    & - \\

    \midrule[1pt]
    \hspace{0.4cm} \textit{Multi-modal FSAR} \\

    CMMN \cite{zhang2020mem} & -  &   -    &    -   & -  &   -    &  -     & 78.92  &   -    &  87.72    & 58.92  &   -    & 70.45 \\
    SRPN \cite{wang2021semguide} &   -    &    -   & -  &  -     &    -   &    -   &   86.50    &  93.80    & 95.80  &   61.60  &     72.50  &  76.20 \\
    KP \cite{shi2022knowledge} &  -    &    -   & -  &  -     &    -   &    \underline{62.40}   &   -    &  -     & \underline{99.40}  &     -  &     -  & \textbf{87.40} \\
    MORN \cite{ni2022morn} &   -    &    -   & \underline{71.70}  &  -     &    -   &    -   &   -    &  -     & 97.70  &     -  &     -  & \underline{87.10} \\
    TADRNet \cite{wang2023dual} &   43.00    &    61.10   & 70.20  &  -     &    -   &    -   &   86.70   &  94.30    & 96.40  &     64.30  &     \underline{74.50}  & 78.20 \\
    EANT \cite{jin2023eant} &   -    &    -   & -  &  -     &    -   &    -   &   87.00   &  \underline{94.60}   & 96.20  &   62.50  &    73.10 & 77.20 \\

    \midrule[1pt]
  
    % SAFSAR (ours) &  &  &  
    %               &  &  & 
    %               &  &  & 
    %               &  &  & \\

    SAFSAR (ours) &  \textbf{71.26}  & \textbf{74.93} & \textbf{76.75}
                    & \textbf{59.23} & \textbf{61.49} &  \textbf{63.68} 
                    & \textbf{98.30} & \textbf{99.31} &  \textbf{99.54} 
                    & \textbf{77.38} & \textbf{83.74} & 85.63  \\

    \bottomrule[1pt]
    
    \end{tabular}%
}
  \caption{Comparison to the state-of-the-art FSAR methods on the meta-testing set of SSv2-Full, SSv2-Small, UCF101, and HMDB51. The experiments are conducted under 5-way 1-shot, 3-shot, and 5-shot settings. ``-" indicates the result is not available for this setting in that paper. Numbers in bold refer to the best results and numbers underlined refer to the second best results.}
  \label{tab:sota}
  
\end{table*}

\begin{table}[htbp]
% \vspace{-0.3em}
% \setlength{\belowcaptionskip}{-16pt} 

  \centering
  \resizebox{0.85\columnwidth}{!}{
    \begin{tabular}{lccc}

    \toprule[1pt]

    \multicolumn{1}{c}{\multirow{2}[0]{*}{Methods}} & \multicolumn{3}{c}{Epic-kitchens}\\

    \cmidrule[1pt]{2-4}
    
    & 1-shot & 3-shot & 5-shot \\

    \midrule[1pt] 

    OTAM &  46.00  & 53.90  & 56.30 \\

    TRX &  43.40  & 53.50  &  58.90 \\

    HyRSM &  \underline{47.40}  & \underline{56.40}  & \textbf{59.80}  \\

    SAFSAR (ours) & \textbf{54.02} & \textbf{58.03} & \underline{59.31}  \\

    \bottomrule[1pt]
    \end{tabular}
    }
  \caption{Results on EPIC-KITCHENS under 5-way 1-shot, 3-shot, and 5-shot settings.}
  \label{tab:epickit_results}
  % \vspace{-4em}
\end{table}

\subsection{The SAFSAR model}
\label{sec:model}
The overall architecture of the SAFSAR model is shown in \cref{fig:model}. We first sample $T$ frames uniformly from each video to extract spatio-temporal features using VideoMAE \cite{tong2022vmae}. Then, the support video features and their corresponding label sentences are processed by the multi-modality fusion (MM Fusion) module to obtain semantic-aware support features, while query video features remain untouched. These features are subsequently input to a task-specific learning module (TLM) to reinforce the interactive cues across videos, getting the final representations. Finally, a cosine similarity distance is computed between query and support final representations to classify the query video.

\vspace{-6pt}
\subsubsection{Spatio-temporal Feature Extraction}
\vspace{-2pt}
% \myparagraph{Spatio-temporal Feature Extraction}
%
Common approaches to FSAR rely on per-frame (2D) feature extraction and later modeling of the temporal dependencies of the extracted spatial features. For instance, in HyRSM \cite{wang2022hyrsm}, they first extract features for each frame and then design a self-attention module operating on the frame-level features to capture long-range temporal dependencies. Arguably, such representations might be suboptimal in capturing the relevant spatio-temporal features for the classification task. In our SAFSAR approach, we propose instead to rely directly on spatio-temporal representations extracted using VideoMAE \cite{tong2022vmae}. If we denote $\Phi_{\mathit{VMAE}}(\cdot)$ as the feature extractor for VideoMAE, we can compute video-based (3D) representations for an input video $\bm x$ 
% of length $T$
to obtain a $d$-dimensional spatio-temporal features:
% \vspace{-1pt}
    \begin{equation}
        \bm f = \Phi_{\mathit{VMAE}}(\bm x),\quad \bm f\in\mathbb{R}^{d}.
    \end{equation}
Then, we follow ProtoNet \cite{snell2017proto} and use averaged video features in the same category as the class prototypes for the support set. Specifically, for a given episode with support set $\mathcal S_e$ we compute:
    \begin{equation}
        \bar{\bm f}_c = \frac{1}{K} \sum_{j\in\mathcal{S}_e} \1{y_j=c} \Phi_{\mathit{VMAE}}(\bm x_j), \quad c \in \mathcal{C}_e,
        \label{eq:prototype}
    \end{equation}
    where $\mathcal{C}_e\subseteq\{1,\ldots,C\}$ is the subset of classes represented in the support set $\mathcal{S}_e$, and where $\1{\cdot}$ is a $0/1$ indicator function taking the value $1$ if the argument is true.
These class prototypes $\bar{\bm f}_c$ are then input to the following modules.

\vspace{-6pt}
\subsubsection{Multi-modality Fusion Module} 
\vspace{-2pt}
% \myparagraph{Multi-modality Fusion Module} 
In this module, the objective is to integrate textual semantics into the visual features. The previous attempts mainly extract embeddings of the label keywords \cite{jin2023eant, wang2023dual,wang2021semguide,wang2021semguide}, \eg, \verb'diving', \verb'brush hair', \verb'golf', \etc using Word2Vec \cite{mikolov2013word2vec} or utilize handcrafted sentence templates \cite{ni2022morn,shi2022knowledge}, \eg, \textit{a video of \{action\}}, as input to a pre-trained text encoder to extract textual semantics. However, employing these strategies does not exploit the full capacity of the pre-trained language models. In our work, we use more detailed and elaborated descriptions provided by \cite{chen2021rehearsal} to better capture nuanced aspects of actions, and enable more meaningful alignment and fusion of both modalities. The elaborated description in \cite{chen2021rehearsal} extends from the class names to a comprehensive definition of the action classes. For example, \verb'Archery' can be expanded as \textit{shooting with a bow and arrows, especially at a target as a sport}, and \verb'LongJump' can be extended as \textit{an athletic event in which competitors jump as far as possible along the ground in one leap}. With the inclusion of these descriptive phrases of the actions, we tap into the advantages of advanced language models to extract more enhanced and discriminative textual semantics.    

As shown in the middle part of the \cref{fig:model}, we first follow the rules in BERT \cite{devlin2018bert} to tokenize the label description sentence into word tokens $\tau \in \mathbb{R}^{L}$, where $L$ is the number of words in the description sentence. Then, we pass the word tokens through the frozen pre-trained BERT-Base model to get the textual features $\bm s \in \mathbb{R}^{L \times d}$ where $d$ is the dimension of the textual features. Next, we concatenate the class prototype $\bar{\bm f}_{c}$ as an extra token to the textual features. The concatenation is then input to the Fusion Transformer module (Fusion TF) which consists of $\ell$ layers of Transformer \cite{vaswani2017tf}. The semantic-aware visual features are then computed as:
\begin{equation}
    [\bm f^*_{c}, \bm s^{*}] = \Phi_\mathit{FTF}([\bar{\bm f}_{c}, \bm s]),\quad \bm s = \Phi_\mathit{BERT}(\tau),
\label{eq:fusion_tf}
\end{equation}
where $\Phi_\mathit{BERT}(\cdot)$ refers to the frozen BERT-Base model, $[\bar{\bm f}_{c}, \bm s]$ represents the concatenation of the class prototype and the textual features, $\Phi_\mathit{FTF}(\cdot)$ denotes the Fusion Transformer module, which adaptively integrates the textual semantics into the visual features, $\bm s^*$ is the output sentence features of the Multi-modal Fusion module and $\bm f^{*}_c$ denotes the semantic-aware support features for class $c$. This paradigm provides a natural mechanism for encouraging the visual encoder to extract semantically consistent features
% aligning visual and textual features 
as the goal is
%in the end we are 
to match the semantic-unaware query video features to semantic-aware support video representations and also facilitate inter-modal feature integration to create coherent representations.
%

% \vspace{-8pt}
\subsubsection{Task-specific Learning Module}
% \vspace{-2pt}
% \myparagraph{Task-specific Learning Module} 
Learning the features of each individual video independently can result into suboptimal performance in FSAR. As evidenced in previous studies, such an approach is often prone to overfitting to irrelevant information and might deteriorate the generalizability of the model to unseen classes \cite{li2019taskrelated, hou2019crosstf}. To address these concerns, we introduce the Task-specific Learning module (TLM), comprised of one Transformer \cite{vaswani2017tf} layer, whose purpose is to augment the video features via cross-video interactions, in order to yield adapted and discriminative representations tailored towards solving the current classification task at each episode. More precisely, we take the semantic-aware support representations and the untouched query features as tokens and apply the TLM on top of them to retrieve the final representations. We denote it as 
    \begin{equation}
        [\{\tilde{\bm f}_{c}\}_{c\in\mathcal{C}_e}, \tilde{\bm f}_{q_e}] = \Phi_\mathit{TLM}\big([\{\bm f^*_{c}\}_{c\in\mathcal{C}_e}, \bm f_{q_e}]\big)
    \end{equation}
    where $\tilde{\bm f}_{c},\, c\in\mathcal{C}_e$ and $\bm f_{q_e}$ represent the final representations of the support videos and the query video, respectively, and $\Phi_\mathit{TLM}(\cdot)$ denotes the Task-specific Learning module.

\subsubsection{The Distance Function}
% \myparagraph{The Distance Function} 
Unlike other methods based on 2D feature extractors that rely on the design of complex distance functions for temporal alignment, and since our approach is inherently 3D, we use instead the cosine similarity to compute the distance between the final representations of the query video and those of the support videos. The cosine similarity between two feature vectors $\bm f_1$ and $\bm f_2$ is defined as:
    \begin{equation}
        \cos(\bm f_1,\bm f_2) = \frac{ \langle \bm f_1,\bm f_2\rangle}{\Vert\bm f_1\rVert_2  \lVert\bm f_2\rVert_2},
    \end{equation}
    where $ \langle \cdot,\cdot \rangle$ denotes inner product.
    Then, we apply softmax to the cosine similarity scores, resulting in the probability of predicting the query video $q_e$ to be a certain action class $c$. Specifically,
    \begin{equation}
        p_c = \frac{\exp ( \cos(\tilde{\bm f}_c,\tilde{\bm f}_{q_e})) }{\sum_{j\in\mathcal{C}_e} \exp( \cos(\tilde{\bm f}_j, \tilde{\bm f}_{q_e}) ) }.
    \end{equation}

\subsection{The Training Losses}
% \myparagraph{Training Loss} 
We use the cross-entropy loss to train the classifier for the correct class assignment:
    \begin{equation}
        L_1 = - \sum_{c\in\mathcal{C}_e} \1{y_{q_e}=c} \log p_c.
    \end{equation}
Additionally, in order to regularize the representations and improve generalization, we include a global classification loss into the training. Specifically, we pass the support and query video features extracted by VideoMAE
% , namely $f_n^k, \ n=1,\cdots,N, \ k=1,\cdots,K$ and $f_i$, 
through one linear layer 
% $g(\cdot):\mathbb{R}^{T\times d}\mapsto\mathbb{R}^C$ 
$\bm W \in \mathbb{R}^{C \times d}$
and then apply a cross-entropy loss to supervise the classification against the total action categories $C$.
% in the training set
We denote this loss as
\begin{equation}\begin{split}
    L_2 = - \frac{1}{NK}\sum_{j\in\mathcal{S}_e} \sum_{c=1}^{C}  \1{y_j=c} \log \sigma_c\big(\bm W \bm f_j \big)\\
    - \sum_{c=1}^{C}  \1{y_{q_e}=c} \log \sigma_c\big(\bm W \bm f_{q_e} \big)
\end{split}
\end{equation}
where we use $\sigma_c(\bm v)$ to denote the $c$-th component of the softmax operation on vector $\bm v$ (\ie, $\sigma_c(\bm v) = \exp{(v_c)}/\sum_j \exp(v_j)$).

Finally, the total loss for training SAFSAR can be expressed as the weighted sum of 
the cosine similarity classification loss and the global classification loss, 
\begin{equation}
    L = L_1 + \lambda L_2
\end{equation}
where $\lambda$ is a weighting factor.

\begin{table*}[htbp]
\setlength{\belowcaptionskip}{-0.35cm} 
  \centering
  \resizebox{0.75\linewidth}{!}{
    \begin{tabular}{@{}crr|rr|rr|rr@{}}

    \toprule[1pt]

    \multicolumn{1}{c}{\multirow{2}[0]{*}{Number of Frames $T$}} & \multicolumn{2}{c}{SSv2-Full} & \multicolumn{2}{c}{SSv2-Small} & \multicolumn{2}{c}{UCF101} & \multicolumn{2}{c}{HMDB51} \\

    \cmidrule[1pt]{2-9}
    
    & 1-shot & 3-shot & 1-shot & 3-shot & 1-shot & 3-shot & 1-shot & 3-shot \\

    \midrule[1pt] 

    8 frames &  71.26  & 74.93  
            & 59.23 & 61.49 
            & 98.30 & 99.31
            & 77.38 & 83.74 \\

    16 frames & \textbf{74.66} & \textbf{78.72}  
            & \textbf{60.69} & \textbf{64.88}
            & \textbf{99.23} & \textbf{99.57}
            & \textbf{78.38} & \textbf{85.71} \\

    \bottomrule[1pt]
    \end{tabular}
    }
  \caption{Comparison between the performance of sampling 8 frames and 16 frames under 5-way 1-shot and 3-shot settings.
  % for SSv2-Full, SSv2-Small, UCF101, and HMDB51.
  }
  \label{tab:ab_nf}
\end{table*}

%-------------------------------------------------------------------------
\section{Experiments}
\label{sec:exp}
\vspace{-5pt}

    \myparagraph{Datasets} We evaluate our method on five few-shot action recognition benchmarks, namely HMDB51 \cite{kuehne2011hmdb51}, SSv2-Full \cite{goyal2017ssv2}, SSv2-Small \cite{goyal2017ssv2}, UCF101 \cite{soomro2012ucf101}, and EPIC-KITCHENS \cite{Damen2018epic}. We do not evaluate on Kinetics-100 \cite{carreira2017kinetics} because our feature extractor was pre-trained on the Kinetics dataset. We adopt the few-shot splits from OTAM \cite{cao2020otam} and CMN \cite{zhu2018cmn} for SSv2-Full and SSv2-Small, respectively, where SSv2-Full contains $10\times$ more videos per action category in the training set. They both consist of 64, 12, and 24 classes for training, validation, and testing. For HMDB51 and UCF101, we use the few-shot splits from ARN \cite{zhang2020arn}. HMDB51 contains 31, 10, and 10 classes while UCF101 contains 70, 10, and 21 classes for training, validation, and testing, respectively. EPIC-KITCHENS is a large-scale egocentric video dataset that includes various actions in kitchens. We use the splits from HyRSM \cite{wang2022hyrsm} which contains 60 and 20 classes for training and testing.

    \myparagraph{Implementation Details} Experiments are conducted using the 5-way $K$-shot ($K \in \{ 1,3,5 \}$) settings. 
    %
%    {\color{blue} We use VideoMAE-Base \cite{tong2022vmae} as the feature extractor which is initialized with Kinetics-710 pre-trained weights \cite{wang2022internvideo}. }
    %
    We use VideoMAE-Base \cite{tong2022vmae} as the feature extractor which is initialized with Kinectis-400 pre-trained weights from \cite{tong2022vmae} for SSv2-Full and SSv2-Small. For all the other benchmarks, we initialize it with Kinetics-710 pre-trained weights from \cite{wang2022internvideo}.
    We uniformly sample $T=8$ frames as in previous methods for a fair comparison. For the text encoder, we utilize BERT-Base \cite{devlin2018bert} initialized with weights provided by \cite{chen2023valor}. In terms of the label descriptive sentences, we use the elaborate descriptions in \cite{chen2021rehearsal} for HMDB51 and UCF101. For SSv2, we directly use the class names as they already provide detailed descriptions for each action class. For EPIC-KITCHENS, we use OpenAI's ChatGPT-3.5 to generate elaborative descriptions based on the class names. During training, we only freeze the patch embedding layer of VideoMAE-Base and the entire 12 layers of BERT-Base and use Adam \cite{kingma2014adam} optimizer to train the model. Basic data augmentation like random cropping, flipping, and color jittering is applied. During inference, we report average accuracy over $E=10,000$ episodes randomly sampled from the test set. 
    % For $K > 1$ settings, we follow ProtoNet \cite{snell2017proto} to get the prototype per category by averaging the features of the support videos and classify the query videos against the prototypes.

\subsection{Comparison with State-of-the-art Methods}
\label{sec:sota}
    In this section, we evaluate the efficacy of our proposed SAFSAR by comparing its performance against existing state-of-the-art (SoTA) approaches on several challenging few-shot action recognition benchmarks. As outlined in \cref{tab:sota}, we order the methods by the publication year and categorize the SoTA methods into two groups, unimodal FSAR and multi-modal FSAR where the latter group uses text as additional information in their models. Remarkably, most research efforts in the multi-modal FSAR group focus on UCF101 and HMDB51, whereas fewer studies examine SSv2. Note that SSv2 is a different type of dataset as it is primarily a motion-centric dataset which emphasizes temporal modeling of the actions while UCF101 and HMDB51 are scene-centric datasets which tend to stress on scene understanding. Moreover, EPIC-KITCHENS is another challenging dataset due to its motion-centric nature and videos captured from a first-person perspective. In addition, this dataset exclusively involves actions performed within a kitchen setting, encompassing a lot of fine-grained classes that demand a comprehensive understanding of the actions. In our experiments, we not only evaluate our proposed SAFSAR in commonly used UCF101 and HMDB51, but also validate it in SSv2 and EPIC-KITCHENS. Our exceptional results verify the effectiveness and robustness of our approach in various types of datasets. 
    
    As shown in \cref{tab:sota} and \cref{tab:epickit_results}, we observe that our proposed SAFSAR achieves remarkable performance and consistently exhibits substantial advantages over other SoTA methods, especially under the challenging 1-shot and 3-shot settings. Concretely, SAFSAR provides around 11\%, 14\%, 7\%, 8\%, and 7\% absolute improvement in SSv2-Full, SSv2-Small, UCF101, HMDB51, and EPIC-KITCHENS respectively, under the most challenging 5-way 1-shot scenario. Additionally, SAFSAR also unequivocally outpaces the other methods by a large margin under 3-shot setting.
    % and 5-shot settings. 
    %{\color{blue} Exceptions are observed for HMDB51 and EPIC-KITCHENS under the 5-shot setup. This drawback could stem from the choice of the number of layers in Fusion TF module. A more detailed discussion can be found in \cref{sec:ablation}. }
    %
    Turning to the 5-shot scenario, SAFSAR continues to outperform the other methods across the majority of benchmarks. While on HMDB51 and EPIC-KITCHEN, SAFSAR's performance is slightly below the state-of-the-art, it still maintains a competitive edge on SSv2-Full, delivering a noteworthy 5\% improvement.
    Of specical note, our SAFSAR achieves substantial improvements on SSv2-Full and SSv2-Small benchmarks, which provides compelling evidence for the superiority of our approach in modeling temporal relations of the actions. 

\subsection{Analysis on the Properties of SAFSAR}
\label{sec:ablation}
% \vspace{-2pt}

% \subsubsection{Number of Frames}
\myparagraph{Number of Frames} 
    In this section, we conduct a thorough investigation of the effect of sampling different number of frames for SSv2-Full, SSv2-Small, UCF101, and HMDB51 under 5-way 1-shot and 3-shot settings. Throughout these experiments, the only change is that we freeze the first 6 layers along with the patch embedding layer and only finetune the last 6 layers in VideoMAE. Motivated by the fact that VideoMAE was pre-trained using 16 frames per video, we may reduce computation cost by fixing early layers, making it comparable with the computation cost of sampling 8 frames. 
    %{\color{blue} The results in \cref{tab:ab_nf} reveal that sampling 16 frames is consistently better than sampling 8 frames. Remarkably, compared with the results of sampling 8 frames, sampling 16 frames brings about 7\%, 3\%, 1\%, and 1\% absolute improvement in SSv2-Full, SSv2-Small, UCF101, and HMDB51, respectively, under the most challenging 5-way 1-shot scenario. These positive effects carry into the 3-shot setting, where an average increase of approximately ?\% can be seen across all four benchmarks.}
    The results in \cref{tab:ab_nf} reveal that sampling 16 frames is consistently better than sampling 8 frames. Remarkably, compared with the results of sampling 8 frames, sampling 16 frames brings about 4\%, 3\%, 0.26\%, and 2\% absolute improvement in SSv2-Full, SSv2-Small, UCF101, and HMDB51, respectively, under the challenging 5-way 3-shot scenario. These positive effects also demonstrate in the 1-shot setting, where an average increase of approximately 1\% can be seen across all four benchmarks.
    The superiority of sampling 16 frames suggests that our proposed SAFSAR is capable of leveraging additional sampled frames to derive richer and better discriminative features, surpassing previous methods. For instance, we notice that although employing increased number of frames generally leads to higher recognition accuracy, some algorithms, such as HyRSM \cite{wang2022hyrsm}, exhibit saturated performance beyond 8 frames as revealed in their ablation analysis. Our findings imply that SAFSAR is capable of making full use of extra frames to achieve superior performance, confirming its strength in accurately encoding temporal relations.

\begin{table}[htbp]
  \centering
  \setlength{\belowcaptionskip}{-0.3cm} 
  
  \resizebox{\linewidth}{!}{
    \begin{tabular}{@{}crr|rr@{}}
    \toprule[1pt]
    
    \multicolumn{1}{c}{\multirow{2}[0]{*}{Number of Layers $l$}} & \multicolumn{2}{c}{SSv2-Small} & \multicolumn{2}{c}{HMDB51} \\

    \cmidrule[1pt]{2-5}
    
    & 1-shot & 3-shot & 1-shot & 3-shot \\

    \midrule[1pt] 

    1  &   57.48 & 61.38   & 77.33 & 83.36 \\
    
    2  &   \textbf{59.23} & \textbf{61.49}   & \textbf{77.38} & \textbf{83.74} \\
   
    3  &   56.73 & 60.83   & 77.31 & 82.69 \\

    \bottomrule[1pt]
    \end{tabular}
}
    \caption{Results of using different $l$ in Fusion FT module under 5-way 1-shot and 3-shot settings.}
  % \caption{Comparison between the performance of using different number of layers of Transformer in Fusion FT module under 5-way 1-shot and 3-shot settings for SSv2-Small and HMDB51.}
  \label{tab:ab_fusion_layers}
\end{table}

% \subsubsection{Number of Layers in the Fusion TF Module}
% \label{sec:ab_num_layers}
\myparagraph{Number of Layers in the Fusion TF Module} 
    As described in \cref{sec:model}, the Fusion TF module consists of $l$ layers of Transformer \cite{vaswani2017tf}. In this section, we study the impact of varying $l$ on the performance of SAFSAR in the SSv2-Small and HMDB51 benchmarks, considering both 5-way 1-shot and 3-shot scenarios. 
%    {\color{blue} According to \cref{tab:ab_fusion_layers}, two notable trends arise: (1) For 1-shot, $l=2$ yields the highest scores; and (2) For 3-shot, $l=3$ produces the peak scores. The reasons behind these observations can be attributed to an increased variations in the visual descriptions associated with larger number of shots, leading to a need for additional layers of the Transformer block in Fusion TF module to effectively handle larger scope of visual cues. While introducing more layers of Transformer in Fusion TF module gains higher accuracy in many-shot settings, we observe that the improvement is limited as it only brings 0.52\% and 0.15\% improvement for SSv2-Small and HMDB51 under 3-shot setup, respectively. Therefore, selecting $l=2$ is optimal for balancing accuracy and computational efficiency which is also the default setting in all other experiments unless otherwise noted. This choice also explains the slightly lower performance reported in \cref{sec:sota} for HMDB51 and EPIC-KITCHENS under 5-shot setting as larger shots might require additional layers in Fusion TF module to better capture and integrate the visual and textual semantics.}
    %
    As observed in \cref{tab:ab_fusion_layers}, it appears that setting $l=2$ yields the best scores under both scenarios across both benchmarks. Therefore, we opt to set $l=2$ as the default setting in all other experiments unless otherwise noted.

% \subsubsection{Analysis of the Proposed Modules}
\myparagraph{Ablation of the Proposed Modules} 
    We summarize the effects of incorporating MM Fusion module and TLM in our proposed SAFSAR in \cref{tab:ab_modules}. The first row serves as a baseline where neither of the module is used. We conduct the experiments in SSv2-Small and SSv2-Full under 5-way 1-shot and 3-shot settings. Particularly, we notice significant improvements when using both modules together, achieving on average about 15\% increase over the baseline for 1-shot setting. Even only using the MM Fusion module alone, it brings appreciable 4\% and 9\% improvement for SSv2-Small and SSv2-Full, respectively, under 1-shot scenario. Similar trend can be observed for the 3-shot setting. Overall, the results confirm the effectiveness of our proposed modules.

    % HyRSM also only used one dataset, SSv2-Full, and conduct the experiments under 1-shot and 5-shot settings. So I guess we can also just choose one dataset and do the experiments under two different settings where I chose SSv2-Small and 1,3-shot.

    \vspace{-2pt}
    
    \begin{table}[htbp]
    \setlength{\belowcaptionskip}{-0.4cm} 
      \centering
      \resizebox{\linewidth}{!}{
        \begin{tabular}{@{}cccc|cc@{}}
        \toprule[1pt]

        \multicolumn{1}{c}{\multirow{2}[0]{*}{MM Fusion module}} & \multicolumn{1}{c}{\multirow{2}[0]{*}{TLM}} & \multicolumn{2}{c}{SSv2-Small} & \multicolumn{2}{c}{SSv2-Full} \\
        
        % MM Fusion module & TLM & 1-shot & 3-shot \\
        
        \cmidrule[1pt]{3-6}
        
        & & 1-shot & 3-shot & 1-shot & 3-shot \\

        \midrule[1pt]

        &                               &  37.87  & 46.00 & 61.18 & 70.96 \\
          
        &\CheckmarkBold                 &  38.88  & 49.20 & 59.30 & 66.51 \\
          
        \CheckmarkBold&                 &  41.14  & 48.19 & 70.51 & 73.94 \\
  
        \CheckmarkBold&\CheckmarkBold   &  \textbf{59.23}  & \textbf{61.49} & \textbf{71.26} & \textbf{74.93} \\

        \bottomrule[1pt]
        \end{tabular}
    }
      \caption{Ablation study on incorporating MM Fusion module and TLM under 5-way 1-shot and 3-shot settings.
      % in SSv2-Small and SSv2-Full.
      }
      \label{tab:ab_modules}
    \end{table}

%{\color{blue}{\myparagraph{Analysis of Computational Efficiency} We run one episode under 5-way 1-shot setting on one NVIDIA RTX A5000 GPU and present the computational efficiency analysis in \cref{tab:computation_analysis}. We observe that TRX has the fewest parameters but incurs the highest computational load, aligning with our rationale that intricate distance measures can lead to increased computational requirements. Conversely, while our model has more parameters due to employing a 3D feature extractor and incorporating a text semantics module which requires an additional language model, it exhibits lower computational demands than TRX thanks to our use of a simple cosine similarity distance function. Comparing our model with the other methods, our model does require more parameters and computations; however, this investment is rewarded by significant performance gains. For example, SAFSAR brings around 15\% and 19\% increase in SSv2-Full and SSv2-Small, respectively.}}

\myparagraph{Analysis of Computational Efficiency} 
    We run one episode under 5-way 1-shot setting on one NVIDIA RTX A5000 GPU and present the computational efficiency analysis in \cref{tab:computation_analysis}. We observe that SAFSAR has more parameters compared to the other methods. The increase is attributed to our utilization of a 3D feature extractor and the incorporation of a language model which constitutes more than half of the total parameter count. Consequently, this architecture choice results in an elevated computational load. While our model does require more parameters and computations, this investment is rewarded by substantial performance gains. For example, SAFSAR demonstrates an average increase of approximately 9\% in both SSv2-Full and SSv2-Small when compared to the other methods. Additionally, despite our model having approximately five times the number of parameters as the other methods, the computational load is only about two times higher. This observation suggests that the computational requirements of our model do not increase linearly with the number of parameters, as demonstrated in the last column of \cref{tab:computation_analysis}. These findings highlight that the computational complexity of our model is not as severe as it might initially appear based solely on the parameter count.
    % In future work, we will focus on strategies to reduce the computational demands while maintaining the current level of performance achieved.

\myparagraph{Analysis of Semantic Consistency} 
    In this section, we qualitatively analyze the semantic consistency in the features extracted by our SAFSAR. We compare the t-SNE \cite{van2008tsne} visualizations of the features extracted with or without the MM Fusion module for five randomly selected actions in HMDB51 and UCF101. As illustrated in \cref{fig:tsne_plot}, we observe that after incorporating the MM Fusion module, the boundaries among different classes become more distinct and expanded. Simultaneously, the features within the same class display increased compactness. This observation signifies our SAFSAR effectively introduces semantic consistency into the features.

\begin{table}[htbp]
\setlength{\belowcaptionskip}{-0.65cm} 
  \centering
  \resizebox{0.9\linewidth}{!}{
    \begin{tabular}{lccc}

    \toprule[1pt]

    Methods & Params (M) & TFLOPs & TFLOPs / Params \\ 

    \midrule[1pt] 

    TRX \cite{perrett2021trx} &  \textbf{47.1}  & 0.340  & 0.007 \\

    HyRSM \cite{wang2022hyrsm} & 65.6 & \textbf{0.333}  & 0.005 \\

    MASTAF (R3D) \cite{zhang2023mastaf} & 48.7 & 0.400 & 0.008 \\

    SAFSAR (ours) & 245.4 & 0.805  & \textbf{0.003} \\
    
    \bottomrule[1pt]
    \end{tabular}
    }
  \caption{Computational efficiency analysis for 5-way 1-shot.}
  \label{tab:computation_analysis}
\end{table}

% ======================================= tsne hmdb
\begin{figure}
    \centering
    \setlength{\belowcaptionskip}{-0.5cm} 

    \includegraphics[trim=0cm 0cm 0cm 0cm, clip, width=0.47\textwidth]{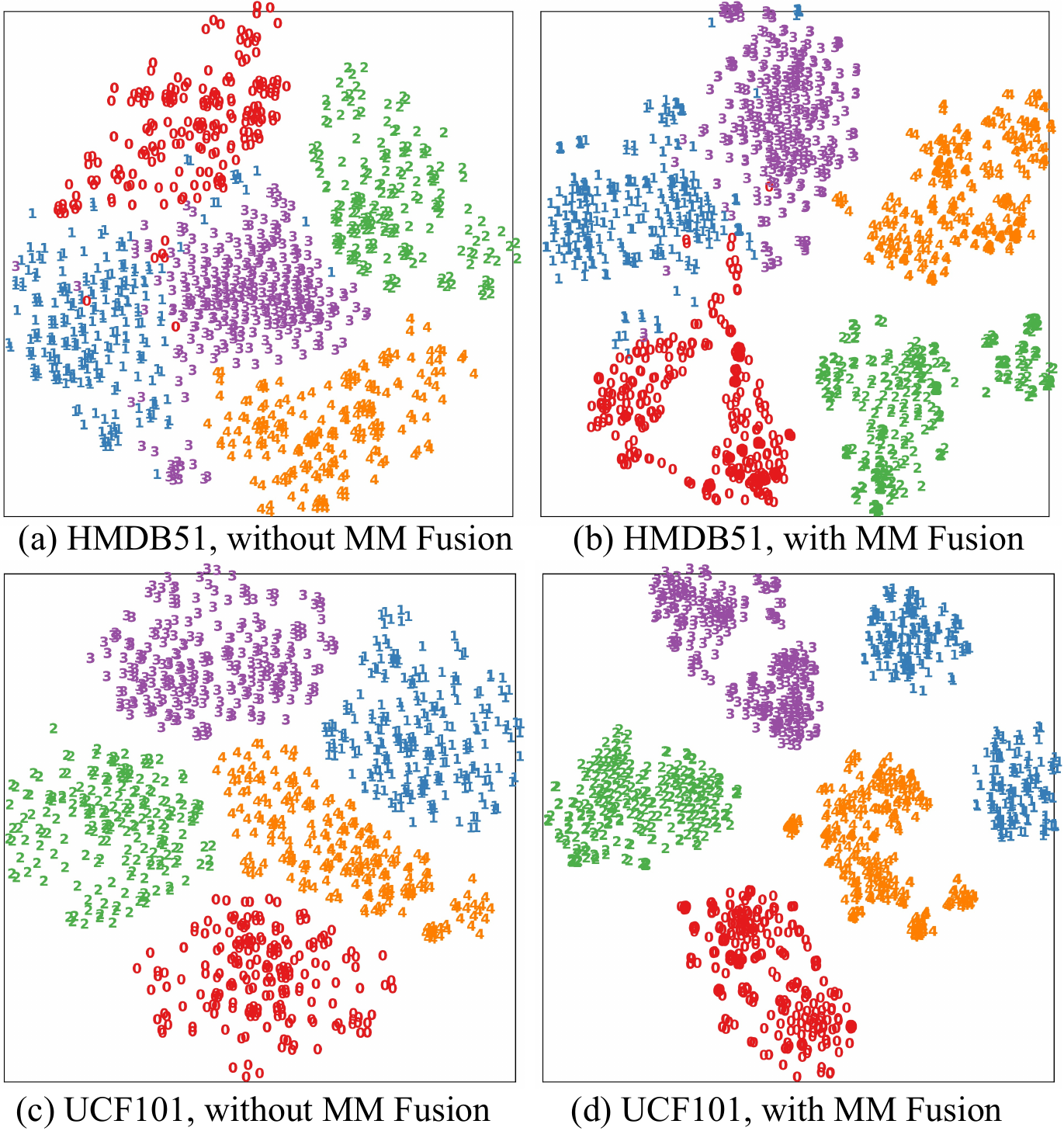}
    
    \caption{t-SNE plots of the features extracted without or with MM Fusion module in the HMDB51 and UCF101 testing set.}
    
    \label{fig:tsne_plot}
\end{figure}
% ======================================= tsne hmdb

% ======================================= tsne ucf
% \begin{figure}
%     \centering
%     % \setlength{\abovecaptionskip}{0.1cm}
%     \setlength{\belowcaptionskip}{-0.25cm}
    
%     % \includegraphics[trim=0cm 0cm 0cm 0cm, clip, width=0.8\textwidth]{figs/model+mmFusion.pdf}
%     \includegraphics[trim=0cm 0cm 0cm 0cm, clip, width=0.475\textwidth]{figs/ucf_combined.pdf}
    
%     \caption{t-SNE visualization of the features extracted by SAFSAR for 5 actions in the UCF101 testing set. \textbf{Left:} Without MM Fusion module. \textbf{Right:} With MM Fusion module.}
    
%     \label{fig:tsne_ucf}
% \end{figure}
% ======================================= tsne ucf

%-------------------------------------------------------------------------
\vspace{2pt}
\section{Conclusion}
\vspace{-4pt}
\label{sec:conclusions}
    In this work, we proposed a novel Semantic-aware Few-shot Action Recognition model (SAFSAR) for few-shot action recognition. It was designed to be simple yet effective by leveraging a 3D feature extractor, VideoMAE, and employing an efficient transformer-based multi-modality fusion module for adaptively integrating textual semantics into the video representations. We conducted extensive experiments to evaluate the performance on five few-shot action recognition benchmarks, including SSv2-Full, SSv2-Small, UCF101, HMDB51, and EPIC-KITCHENS. The experimental results demonstrated the efficacy of our proposed SAFSAR model by improving upon existing methods and achieving state-of-the-art performance.

\vspace{-2pt}
\section{Acknowledgements}
\vspace{-2pt}
The authors thank Kaleab Kinfu and Carolina Pacheco for their
valuable feedbacks. This work was supported by a Discovery Challenge Award from Johns Hopkins Discovery Fund Program.

%%%%%%%%% REFERENCES
{\small
\balance
\bibliographystyle{ieee_fullname}
\bibliography{SAFSAR}
% \bibliography{egbib}
}

\end{document}